# An Explainable AI System for the Diagnosis of High Dimensional Biomedical Data

Alfred Ultsch[1], Jörg Hoffmann[2], Maximilian Röhnert, Malte Von Bonin[3], Uta Oelschlägel[3], Cornelia Brendel[2], and Michael C. Thrun[1,2]

1) Databionics, Computer Science, Philipps-Universität Marburg, Hans-Meerwein-Straße 6 D-35032 Marburg.
2) J Department of Hematology, Oncology and Immunology, Philipps-University, Baldinger Str., D-35032 Marburg.
3)Medizinische Klinik und Poliklinik I Bereich Innere Medizin / Hämatologie und Onkologie, Universitätsklinikum Carl Gustav Carus an der Technischen Universität Dresden, Fetscherstraße 74, 01307 Dresden.

CORRESPONDING AUTHOR
E-mail: mthrun@mathematik.uni-marburg.de
ORCID 0000-0001-9542-5543

## ABSTRACT

Typical state of the art flow cytometry data samples consists of measures typically 10 to 30 features of more than 100.000 cell "events". AI systems are able to diagnose such data with almost the same accuracy as human experts. However, there is one central challenge in such systems: their decisions have far-reaching consequences for the health and life of people. Therefore, the decisions of AI systems need to be understandable and justifiable by humans. In this work, we present a novel explainable AI method, called ALPODS, which is able to classify (diagnose) cases based on clusters, i.e., subpopulations, in the high-dimensional data. ALPODS is able to explain its decisions in a form that is understandable to human experts. For the identified subpopulations, fuzzy reasoning rules expressed in the typical language of domain experts are generated. A visualization method based on these rules allows human experts to understand the reasoning used by the AI system. A comparison to a selection of state-of-the-art explainable AI systems shows that ALPODS operates efficiently on known benchmark data and also on everyday routine case data.

## 1. INTRODUCTION

State of the art machine learning (ML) artificial intelligence (AI) algorithms are effectively and efficiently able to diagnose (classify) high-dimensional data sets in modern medicine, see [Keyes et al., 2020] for an overview. In particular for multiparameter flow cytometry data, see [Hu et al., 2019; Zhao et al., 2020]. These are systems that use one set of data, the learning data, to develop (train/learn) an algorithm which is able to classify data that is not part of the training data (i.e. the test or validation data). This is called supervised learning [Murphy, 2012]. Most successful in supervised learning are artificial neuronal networks (ANN) consisting of simple processing units (neurons) organized in interconnected layers (deep learning ANN) [Goodfellow et al., 2016]. Within artificial intelligence, these algorithms are called subsymbolic classifiers [Ultsch, 1998]. Subsymbolic systems are able to perform a task (skill), such as assigning the most suitable diagnosis to a case. However, it is meaningless and impossible to ask a subsymbolic AI system for an explanation or reason for its decisions (e.g., [Tjoa/Guan, 2020]). To circumvent this lack of explainability there has been a recent explosion of work in which a second (post hoc) model ("post-hoc explainer) is created to explain black box machine learning model [Rudin, 2019]. Alternatively, psychological research proposed to us category norms that are derived by recruiting exemplars as representations of knowledge [Kahneman/Miller, 1986]. In machine learning this psychological concept is called prototyping [Sen/Knight, 1995] and well-known in cluster analysis (e.g., [Nakamura/Kehtarnavaz, 1998]) and neural networks (e.g., [Vesanto, 1999; Thrun/Ultsch, 2020b]). Recently, Angelov and Soares proposed to integrate the selection of prototypes in a deep learning network [Angelov/Soares, 2020]. A pre-trained traditional deep learning image classifier (CNN) is combined with a prototype selection process to xDNN [Angelov/Soares, 2020].

In particular, in medicine, explanations and reasons for decisions made by algorithms concerning the health state or treatment options for patients are required by law, e.g., GDPR in the EU[1]. This calls for systems that produce human-understandable knowledge out of the data and base their decisions on this knowledge such that these systems can explain the reason for a particular decision. Such systems are called  "symbolic" or "explainable AI (XAI)" [Adadi/Berrada, 2018].  If such systems aim to represent their reasoning in a form, which is understandable to application domain users, they are called  "knowledge-based" or "Expert Systems" [Hayes-Roth et al., 1983].

Some XAI systems produce classification rules in the form of a set of conditions. If the conditions are fulfilled, a particular diagnosis is derived. For example, a "thrombocyte" can be described with the rule: CD45-, CD42+. The conditions of the rules consist of logical statements on the range of parameters (variables). For example, "CD45-" denotes the condition that the expression of CD45 structures, measured by the flow cytometry device, on a cell's surface is low. In AI systems, the production of diagnostic rules from data sets is a classical approach. Algorithms like CART [Breiman et al., 1984], C4.5 [Salzberg, 1994], RIPPER [W. W. Cohen, 1995] were already developed in the last century. However, these algorithms aimed for optimizing the performance of the classification and not for best understandability of the rules to domain users.

---

[1] European Parliament and Council : General Data Protection Regulation (GDPR), in effect since25 May 2018

One of the essential requirements for human understandability of machine-generated knowledge is simplicity. If a rule comprises too many conditions or if there are to many rules, their meaning is very hard or impossible to understand. In AI the simplicity of the descriptions is a typical quality measure for understandability [Dehuri/Mall, 2006].

This work proposes a symbolic machine learning algorithm (ALPODS) that produces and uses user understandable knowledge for its decisions. The algorithm is tested on a typical ML example and two multiparameter flow cytometry data from the clinical routine. The latter are from two clinical centers (Marburg and Dresden). The problem at hand is to decide the primary origin of a probe: either mostly bone marrow or peripheral blood. The algorithm is compared to rule generating AI systems [Ribeiro et al., 2016] and decision tree rules [Loyola-González et al., 2020], and to recently published rule generating algorithms for flow cytometry [Aghaeepour et al., 2012; O'Neill et al., 2014].

The manuscript is structured as follows. Section two introduces the particular multivariate data challenge posed by the particularities of flow cytometry data. In three subsections XAI algorithms are described which are, in principle, able to tackle the problem at hand. After this review of related work the new algorithm called ALPODS is introduced in section three. Section four describes the evaluation approach and the used datasets. Results are presented in the fifth section. This work finishes with the discussion and conclusion in section six.

## 2. RELATED WORK

Recent review and commentary articles on XAI imply implicitly or explicitly assume, that there is a general trade-off between understandability and accuracy [Rudin, 2019]. For example, Burkart and Huber define that XAI aims for solving an optimization task by minimizing the averaged error overall training instances of supervised machine learning [Burkart/Huber, 2021], p.253. Several types of XAI approaches based on this notion. A model derived in this way is applied to new data to obtain a prediction [Burkart/Huber, 2021] which then should be explained. However, prediction requires only correlation [Holzinger et al., 2020]. This means that the aim is to correlate the model's output (prediction/classification) based on the input, i.e., given data used in the model.

In contrast, understandability requires comprehensible models in the sense that their steps to arrive at a conclusion are interpretable by and explainable to humans [Holzinger et al., 2020]. Such models have causality as an ultimate goal for understanding underlying mechanisms [Holzinger et al., 2019]. Rudin claims that systems with the aim of being understandable to domain experts [T. Miller, 2019] should still perform similar to XAIs which only achieve correlations between input and output of its models [Rudin, 2019].

An overview of selected artificial intelligence algorithms is presented below, which produce diagnostic decisions for multivariate high-dimensional data such as flow cytometry data. Such data typically comprises several cases (patients). For each case a fairly large number of events (cells) ($n \geq 100.000$) are measured. In cytometry for each cell, the expression (presence) of proteins on their surface is measured (variables). The proteins are genetically encoded by cluster of differentiation (CD) genes [Mason, 2002]. For the data, a classification into k classes is given (diagnosis): each case (patient) is classified, typically encoded as a number from $(1, \ldots, k)$. An important issue in the analysis of such data is the automated identification of populations (clusters) in the high dimensional (multivariate) data. Such populations may be either relevant for gating the data, i.e., eliminating unwanted cells or debris [Shapiro, 2005] or, more importantly, the (sub-) populations are relevant for a diagnosis.

The explainable AI (XAI) algorithm "ALPODS" presented here operates as follows: subsets of the events of all cases are calculated. These populations are called subpopulations or clusters. Clusters which are not relevant for the decision are not considered further. For each of the remaining k clusters, an explanation is produced. An explanation for a cluster consists of a number of conditions that are able to select the members of the cluster out of the data set. A standard notation for conditions for the description of cell types in flow cytometry is the plus-minus (+-) notation [Wood et al., 2007]. For example, a cell population that contains intermediate forms of T-cells may be described if the form of this rule: "CD3-, CD4+, + CD8+" [Wood et al., 2007]. The comma in the rule (",") means logical "and" (conjunction) whereas "+" and "-" denote the over- respectively under- presence of the cell surface structures in the described T-cells.

In summary, in this particular data challenge each case is described by many events and each event by several features. AI algorithms are, however, used to explain the diagnosis of a case and not an event. For such problems typically either decision trees or machine learning methods combined with post-hoc explainers are used. Hence, the first subsection introduces conventional supervised [Breiman et al., 1984] and unsupervised decision trees [Loyola-González et al., 2020]. The second subsection introduces a more general concept of combining a machine learning system based on a large number of degenerate decision trees (random forest) with a post-hoc explainers which represents an approach to extract explanations based on a learned model [Lipton, 2018]. A well performing machine learning model (random forests) with a conventional post-hoc explainer based on a linear model with LASSO regularization [Ribeiro et al., 2016] is presented. For this type of approach a variety of alternatives are proposed, see [Lundberg/Lee, 2017].

The third subsection introduces two typical domain specific algorithms used in flow cytometry. The first one called SuperFlowType constructs event hierarchies to provide the best population selection strategies to identify a target population to a desired level of correlation with a clinical outcome, using the simplest possible marker panels[Aghaeepour et al., 2012]. The second one is called "Full Annotation Using Shaped-constrained Trees" (FAUST)[Greene et al., 2021] which delivers descriptions of a large number of populations relevant for a particular disease[Vick et al., 2021].

### 2.1 Supervised and Unsupervised Decision Trees

XAI algorithms often use either explicitly or implicitly decision trees. Decision trees are especially popular for medical diagnosis because they assumed to be easy to comprehend [Ripley, 2007] (chapter 7: tree-structured classifiers). Decision trees consist of a hierarchy of decisions. At each (decision-) node of the tree, a variable (CDxxx) and a threshold t is selected. The two possibilities of explanations based on single features, "CDxxx ≤ t" vs. "CDxxx > t" split the considered data set into two disjunct subsets. For each of these possibilities, descendants in the tree are generated. Different decision tree algorithms use different approaches for the selection of the decision criterion, i.e., the variable CDxxx and threshold t (selection criterion). The construction starts with the complete data set and typically ends if either a descendant node contains only cells of one class (the same diagnosis) or a stop criterion on the size of the remaining subsets are reached (stop criterion). Decision trees are supervised algorithms, i.e., for the selection and stop criterion, the predefined classification (diagnosis) of the cases is required. A popular decision tree algorithm is the classification and regression tree (CART) [Breiman et al., 1984]. CART is used in this work as a baseline of performance. We use the CART implementation in form of the R package "rpart" available on CRAN (https://CRAN.R-project.org/package=rpart) in this work.

Exemplary for the domain of unsupervised decision trees, the algorithms Full Annotation Using Shaped-constrained Trees (FAUST)[Greene et al., 2021] and eUD3.5[Loyola-González et al., 2020] are selected. FAUST [Greene et al., 2021] produces a forest of unsupervised decision trees. FAUST is described in detail in section 2.3. Loyola-González et al. proposed in 2020 an unsupervised decision tree algorithm called eUD3.5 [Loyola-González et al., 2020]. eUD3.5 uses a split criterion based on the silhouette index [Loyola-González et al., 2020]. The silhouette index compares the homogeneity of a cluster with the heterogeneity of other clusters [Rousseeuw, 1987]. In eUD3.5 a node is split only if it's possible descendants have a better split criterion than the best split criterion found so far. This leads to a decision tree that is based on the cluster structures (homogeneity) and not on the diagnosis. A cluster is labeled with a particular diagnosis by the majority of diagnoses of the members in the cluster. In eUD3.5, 100 different trees are generated, their performance is evaluated, and the best performing tree is kept. The user can specify the number of desired leaf nodes (stop criterion). If the algorithm produced more leaf nodes than specified by the user, then leaf nodes are combined using k-means. No open-source code was referenced in [Loyola-González et al., 2020]. A direct request for source-code was unsuccessful. No suitable code for Python, Matlab or R was provided by the authors. Therefore all the results of eUD3.5 are taken from [Loyola-González et al., 2020].

Other alternatives, either do not provide source-code [Dasgupta et al., 2020] or are unable to process n>100.000 events per case, if the run time is limited below 72 hours of computing time [Thrun et al., 2021; Thrun, 2022].

## 2.2 Random Forest with LIME (RF-LIME)

There is a widespread belief that black-box, such models are more accurate than understandable models [Rudin, 2019] Hence, one of the allegedly best-performing machine learning algorithms, random forest (RF), was selected [Fernández Delgado et al., 2014; Wainberg et al., 2016] to investigate this claim. In addition the chosen Open MP implementation of RF (see SI A for details) used here computes results very efficiently which is often a requirement in the medical domain. RF uses many small decision trees, for which the training data of each tree is randomly subsampled from the data [Breiman, 2001]. A random subset of the variables is evaluated at each decision node. The split criterion is usually the same as in CART [Breiman et al., 1984], [Grabmeier/Lambe, 2007]. Many, typically n=500, trees are considered (forest), which produce many individual classifications. The forest's classification is determined as the majority vote of all the tree classifications. For explanations, i.e., rules for the populations, the program LIME [Ribeiro et al., 2016] is used because it is a well-known representative for post-hoc explainers [Burkart/Huber, 2021]. In comparison to alternative post-hoc explainers [Jesus et al., 2021] LIME's produces substantially lower variability of explanations from case to case which is required for understandability. LIME reduced the decision tree to a k-dimensional multivariate regression model.

The user must provide the number k of variables used for the regression. From this model, a set of conditions is extracted. The detailed parameter settings and implementation of LIME is described in supplementary A (SI A).

### 2.3 Comparable XAI Methods used in Flow Cytometry

Two published methods which operate on flow cytometry data and produce descriptions for subpopulations are described for close comparison to our approach: Supervised FlowType/RchyOptimyx (SuperFlowType) [Aghaeepour et al., 2012; O'Neill et al., 2014] and Full Annotation Using Shaped-constrained Trees (FAUST)[Greene et al., 2021]. Both methods claim to produce explainable AI systems.

The FlowType [Aghaeepour et al., 2012; O'Neill et al., 2014] algorithm consist of a brute force approach that exhaustively enumerates all possible population descriptions defined by all CD variables. For each variable CDi, three different situations are considered: CDi+, CDi- and no condition on CDi. The thresholds for the + and – comparisons are calculated by a one-dimensional clustering algorithm using the two "cluster": low range vs. high range of the variable. As decision algorithms for the low/high regions either k-means [Linde et al., 1980] or flowClust [Lo et al., 2009] are used. FlowType generates rules for all possible combinations of conditions in all d variables. For d variables and the three possibilities (not used, + and -) this results in $3^d$ populations. For a typical state of the art flow cytometry data set with d=10, these results in n = $3^{10}$ = 59049 possible cell populations. To reduce this number of cell populations, the RchyOptimyx algorithm is applied [Aghaeepour et al., 2012; O'Neill et al., 2014]. RchyOptimyx applies graph theory, dynamic programing and statistical testing to identify the best subpopulations which are added to a directed acyclic graph (DAG) as a representation of hierarchical decisions. SuperFlowType has been used to evaluate standardized immunological panels [Villanova et al., 2013] and to optimize lymphoma diagnosis [Craig et al., 2014]. In the FlowCAP benchmarking data the application of SuperFlowType resulted in a F-measure of 0.95 [Aghaeepour et al., 2013] and provided statistically significant predictive value in the blinded test set [Aghaeepour et al., 2016]. [O'Neill et al., 2014] claim that SuperFlowType is able, within the memory of a common, workstation (12GB), to analyze 34-marker data [O'Neill et al., 2014]. For the datasets used here the computational load for the application of SuperFlowType was observed to be extreme high (see results chapter).

The FAUST algorithm consists of two phases [Greene et al., 2021]: in the first phase a feature reduction method is applied. All possible combinations of three CD markers are considered. A combination is termed "reasonable", if the Hartigan's dip test [Hartigan/Hartigan, 1985] rejects unimodality of the distributions. All reasonable combinations are added to a decision tree for a particular case (patient), called "experimental unit" in [Greene et al., 2021]. Across all decision trees of all experimental units the distribution of the depth in which a particular marker combination is recorded. Features with shallow depth values are removed from further considerations (feature selection). In the second phase the remaining decision trees are pruned by their frequency of phenotype occurrence across different cases. Each subpopulation in FAUST described in form of a set of CDi+, CDj- conditions.

## 3 ALPODS – THE ALGORITHMIC POPULATION DESCRIPTIONS METHOD

The algorithm called "Algorithmic Population Descriptions" (ALPODS) is constructed as a Bayes decision network in the form of a directed acyclic graph (DAG). The decision network is recursively constructed as follows: first, a variable is selected for the current node $o$ of the DAG For selecting a variable, the conditional dependencies are evaluated using the Simpson Index (S). S represents the expected joint probability that two entities taken from the population represent the same or different types. It is calculated as the conjunction of the probability that an entity belongs to the population and the probability that the entity does not belong to a particular population. This can also be interpreted as the probability that two randomly chosen events do not belong to the same subpopulation. S is biologically inspired as the expected heterozygosity in population genetics, respectively the probability of interspecific encounter [Hurulbert, 1971]. Conditionals dependencies are modeled using the theorem of Bayes [McGrayne, 2011] on the probability distribution of the parent node (prior) and the probability distribution of the decedent node (posterior). Using Bayesian decisions ensures the optimal decision in terms of costs (risk) of a decision [Ruck et al., 1990]. The details of the DAG construction are described as pseudocode in Listing 1.

```
Algorithm   ALPODS
DAG = GrowDAGforALPODS(DAG,CDi,depth, Population,Classification)
Input:
     DAG
     CDi
     depth
     Population
     Classification

 1: if TerminationCondition(depth,size(Population), Classification)  then
 2:     leafLabel = Classify(Population,Classification)                                                    ▷ majority vote
 3:     DAG = AddLeaftoDAG(DAG, leafLabel)                                                      ▷ DAG is complete at this leaf
 4: else                                                                                      ▷ termination not reached
 5:     for each variable CDi do do
 6:         for each plausible condition C do do
 7:             let C(Population) be the subpopulation resulting from
 8:             application of C.
 9:             use the probability distributions PDF(Population) and
10:             PDF(C(Population)) to
11:             calculate a decision on membership using Theorem of Bayes
12:             ⇒ ClassOfSubpopulation
13:             SI(CDi) = SimpsonIndex(Population, C(Population))
14:             if significant(SI(CDi)) then
15:                 AddEdgeToDAG(C(Population),Label= "C on CDi")              ▷ NOTE: edge is only added if it is not already descendant of current DAG
16:                 DAG = GrowDAGforALPODS(DAG,CDi,depth+1, Subpopulation, ClassOfSubpopulation)        ▷ expand the DAG recursively.
17:             end if                                                                            ▷ only for suitable SIs
18:         end for                                                                              ▷ for all conditions
19:     end for                                                                                ▷ for all variables
20: end if                                                                                    ▷ expansion of the DAG

Used Procedures
 1: TerminationCondition()                                          ▷ is true if depth or size exceeds limits or classification contains only one class
 2: AddLeaftoDAG()                                                                    ▷ adds a Leaf node to the DAG
 3: AddEdgeToDAG(Population,Label)                              ▷ adds edge with Label to is only added if it is not already descendant of current DAG
 4: significant()                                                                    ▷ Threshold for the Simpson Index
```

Listing 1: Pseudo Code for the generation of the DAG in ALPODS.

Second, edges are generated and associated with conditional dependencies to descendant nodes, and third, either recursion stops when a stop criterion is fulfilled or the DAG construction is recursively applied to all descendant nodes. Recursion starts with the complete data set as the population. Recursion stops when the class labels are identical for all members of a subpopulation, or the sub population's size is below a predefined fraction of the data, typically one percent. This results in a Bayesian decision network that is able to identify subpopulations within the cells. A comprehensive overview over Bayesian networks in comparison to decision trees can be found in (Freitas 2014). In order to explain the subpopulations, the decisions are summarized into explanations.  Following the idea of fast-and-frugal trees [Luan et al., 2011], the sequence of decisions for a population is simplified into an algorithmic population description: all Bayesian decisions i.e., conditions that use the same marker, are simplified into a single condition (see below).  This describes an interval within the range of this marker.  The relevance of the subpopulation for the diagnosis is assessed by an effect size measure [Wilson/Sherrell, 1993], by default, the absolute value of Cohen's d is used [J. Cohen, 2013]. Computed ABC analysis [Ultsch/Lötsch, 2015] on the effect sizes is recursively applied to select the m most relevant few subpopulations until a number of populations is reached, which lies within the miller optimum m of human understanding, i.e.: m=7+-2 [G. A. Miller, 1956]. The simplification aims to describe a population that is understandable by human experts. The subpopulations' descriptions are presented to the domain expert, who is asked to assign a meaningful name for each of the relevant population.

In the interaction with the domain experts, it became apparent that rules with conditions resulting from an XAI system alone are not sufficient to understand the subpopulation selected by the rule. Therefore in order to address the "meaning" of a certain cell population a matrix of class colored scatterplots called visualization panel (VisPanel), was presented to the clinical experts. The number and layout of the VisPanel can be either determined by the experts in the same way as they are used to in everyday clinical routine. Alternatively, it is determined by the largest absolute probability differences (ProbDiff). ProbDiff is calculated for all pairs (X,Y) of the variables in the data.  For each pair (X,Y) of variables, the probability density of a class C p(X,Y|C) is estimated using smoothed data histograms SDH [Eilers/Goeman, 2004]. With this the absolute probability difference results to: $ProbDiff(X, Y, C)) = |SDH(X, Y|C) - SDH(X, Y|not(C))|$.

The computed ABC analysis[Ultsch/Lötsch, 2015] is applied to the ProbDiffs. The A- set, i.e. the largest few is selected through the computed ABC analysis and the ProbDiffs of set A constitute the VisPanel. The class with the highest class probability (for example, bone marrow in Rule 1) is depicted using red dots on top of the data outside of the class (peripheral blood in the example). This differential population visualization method (DIPOLVIS) allowed the clinical experts to understand all presented populations and to assign meaningful descriptions to each XAI result. In the above case of Rule1, the population one was identified as "myeloid progenitor cells"

For example, for the rule "Rule1: events relevant for bone marrow are in Population 1"SS++ and CD33+ and CD13" for which the VisPanel looks like as shown in Fig. 1. The events of "Population1" are marked in red. Population1 cells occur in bone marrow probes with an average of 43% and in peripheral blood with 5%. The left panel of Fig. 1 reveals that the cell population is located in the area of myeloid cells with high side and forward scatter (SS and FS) apart from lymphocytes and monocytes. The middle panel reveals that the cell population has dim CD45 expression within the myeloid cells. Finally, Population1 could be identified as myeloid progenitor cells (e.g., myocytes) because of weak CD13 and strong CD33 expression (right panel).

Combining the relevant population to an XAI diagnostic system is based on fuzzy reasoning [Mamdani/Assilian, 1975]: the relative fractions of membership for the relevant subpopulations on a selected subset of patient data (extended learning set) is calculated. Fuzzy membership functions are calculated as the posterior probabilities. Figure 2 shows an example of this approach. On the left panel of Figure 2 the probability distributions (PDF) in the variable SS (side scatter) of a particular subpopulation of the data is shown.

The Theorem of Bayes defines the shape of the fuzzy membership function for "SS+ in bone marrow", respectively many("SS","BM"). The decision limit according to the Theorem of Bayes is shown as the black line. This can be used for the calculation of the SI index and as well for the simplification of the description of a particular population as stated above. For example, if another condition on SS for BM is less restrictive, that condition can be omitted. Within the range 2.4 to 3.2 for SS in the example of Figure 2, the membership function is indecisive which expands the diagnostic capabilities of the XAI. The fuzzy conjunction of all single XAI experts decides the classification (diagnosis). This results in a set of XAI experts for each of the populations.

The big advantage of this approach that both, the pro and the contra, for a diagnosis, can be explained to the domain expert in terms that are understandable. For example, an explanation, why a probe contains more peripheral blood than bone marrow looked like: many (thrombocyte aggregates) and few (progenitor B-cells). The functions many() and few() are defined as fuzzy set membership functions [Mamdani/Assilian, 1975].

## 4. MATERIAL AND METHODS

### 4.1 Data Description

All explainable AI algorithms regarded here were tested on one well-known machine learning data set (Iris) and three different Flow Cytometry data sets. The first two consists of samples of either peripheral blood (PB) or bone marrow (BM) from patients without any sign of bone marrow disease at two different health care centers (University Clinic Dresden and University Clinic Marburg). The manual distinction of such samples poses a challenge for the domain expert. The third dataset contains healthy bone marrow samples and leukemia bone marrow samples because diagnosis of leukaemia based on bone marrow samples is a basic task.

The IRIS data serves as a basic test of performance of the algorithms introduced. As stated in the data descripton [Thrun et al., 2022] the flow cytometry data are derived from original diagnostic sample measurements that were obtained to get information on the MRD level in acute myeloid leukemia (AML) (c.f. [Bacigalupo et al., 1992; Muschler et al., 1997]). For clinicians it is an essential question whether bone marrow (BM) samples of patients with haematological diseases are diluted with peripheral blood (pB) in order to assess Minimal Residual Disease status (MRD) in the correct manner. Domain experts (i.e., clinicians) distinguish pB samples from BM samples, and leukemia BM samples versus non-leukemia BM samples based on distributions of biological cell populations by looking at two-dimensional scatter plots, i.e., clear, straightforward patterns in data that have a biological meaning and are visible to the human eye (c.f., [Thrun/Ultsch, 2020a]), which means that the cases of each sample can be used to benchmark machine learning methods (c.f.,[Thrun, 2021]). Three types of assessment of XAI methods are performed by using the data: performance, understandability and processing time.

### 4.1.1 The Iris Data With Gaussian Noise

The Iris data set describes three types of 150 Iris flowers [Anderson, 1935]. The data set consists of 50 samples from each of three species: Iris setosa, Iris virginica, and Iris versicolor. Four features are given for each sample: the lengths and widths of the sepals and lengths and widths of the petals. The class of setosa is well separated from the other classes but the other two classes, called virginica and versicolor, overlap [Setzu et al., 2021]. This presents "a challenge for any sensitive classifier" [Ritter, 2014].

In order to obtain sufficiently large training and testing data sets, Gaussian noise with zero mean, and small variance N(0,s) is added to the data values. As variance ($s^2$) ten percent of the variance in each of the four variables is used. This procedure (jittering) was applied 10 times resulting in 1500 cases that are divided into equal-sized Iris-training and -test sets of n=750 cases each. In order to make sure that the jittered data are structurally equivalent to the original Iris dataset, a random forest classifier was trained with the jittered data. This random forest classifier was able to classify the jittered data with an accuracy of 99% (results not shown), i.e., the jittered data can be used for training and testing XAI algorithms as well as the original iris data.

*4.1.2  Marburg and Dresden data*

Retrospective reanalysis of the flow cytometry data from blood- and bone marrow samples was done according to the guidelines of the local ethics committee. Sample data were acquired with two different flow cytometers: Navios™, Beckman Coulter (Krefeld) for the Marburg data and BD FACSCanto II™, BD Biosciences (Heidelberg) for the Dresden Data. Both measure forward and side scatter and using the same panel of fluorescent antibody clones against the same antigens (CD34 FITC (8G12), CD13 PE (L138), CD7 PerCP-Cy5.5 (M-T701) CD56 APC (NCAM16.2) all BD Bioscience and CD33 PE-Cy7 (D3HL60.251), CD117 AlexaFluor750 (104D2D1), HLA-DR Pacific blue (Immu357), CD45 Krome Orange (J33) all Beckman Coulter).

The Marburg dataset consisted of n = 7 data files (samples) containing event measures from peripheral blood (BP) and n = 7 data files for bone marrow (BM). The Dresden dataset comprised of n = 22 sample files for peripheral blood and n = 22 samples for bone marrow. Each sample file contained more than 100.000 events. The files were randomly subsampled for the training data and combined with a single training data set. The Marburg data contains n= 700.000 events, and the Dresden data set n= 440.000 events. The classifications for the events were derived from the type of the sample (BM or PB). The training data was used in the XAI algorithms to generate populations and corresponding decision rules. Training of XAIs is performed event-based on a balanced sample. Evaluation of performance is performed per patient sample with the measure of accuracy because the number of blood samples and bone marrow samples is balanced. Testing was performed on the sample data files for which a decision (BM of PB) was calculated as a majority vote of the population classifiers.

The exception is FAUST which serves as a feature extraction method. The rule-based explanations generated on the training files are applied to the sample files resulting in cases (sample files) of extracted features of cell counts (rules). These cell counts are used as the input for random forests (SI A for specific algorithm) for which the target is the class of the sample file. For the N=14 sample files of Marburg, a leave-one-out cross-validation is performed and for the other datasets, the n-fold cross-validation is performed on a 50% test data after learning the random forest classifier on the 50% of training files. To provide an easy example for explanations in ALPODS, as another dataset, N=25 healthy bone marrow samples and N=25 leukemia bone marrow samples are used. Detailed descriptions of flow cytometry datasets and availability are described in (Thrun, Hoffman

et al. 2022).

### 4.2 Assessing the Quality of Explainable AI Systems

Explainable AI systems (XAI) deliver a set of n = #c clusters (subpopulations) of a dataset that are presumably relevant for distinguishing a particular diagnosis d from all data which do not fit into the diagnosis class (not (d)). So the first criterion to be applied to an XAI system is to measure how each cluster matches the classification. This is usually measured using accuracy, which is the percentage of elements of a cluster consistent with the diagnosis class [Florkowski, 2008].
The second aspect of XAI systems is the question of understandability. Understandability by itself is a multifactorial issue (see [Langer et al., 1999] for a discussion). However, one of the basic requirements for understandability is the simplicity of an explanation [Dehuri/Mall, 2006]. This can be measured as follows: XAI systems deliver not only a number of clusters #c but also rules that describe the content of the using a number of conditions (#cond). In order to understand the results of an XAI system both, #c and #cond must be in a human-understandable range. According to one of the most often cited papers in psychology, the best memorization and understanding in humans is possible if the number of items presented is around n=7 (Miller Number) [G. A. Miller, 1956]. Consequently, the sizes of results, #c, respectively #cond, of an XAI system are considered not to be understandable if they are either trivial with n < 2 or too complex with n > 14 [Dehuri/Mall, 2006].

### 4.3 Experiments

#### 4.3.1 Randomized Iris Data

The results of eUD3.5 are cited from [Loyola-González et al., 2020].The random forest (RF) was constructed using the training data set (n = 750). This RF was applied to classify the n =750) test data. The accuracy of the RF classifier was evaluated on the test data for 50 rounds of cross-validation. For each case, LIME provided a local explanation through a rule. The case-wise rules were aggregated using the algorithm that generates the facetted heatmap of LIME. Then the rules could be extracted from the x-axis of this heatmap (see Fig.5, SI A). For SuperFlowType, three binary classifications, one iris flower type vs. all others, were constructed (n = 250 cases). Rules were generated by comparing class 1 to the combination of classes 2 and 3, class 2 to 1 and 3 and class 3 to 1 and 2 and applied to the test data set (n =750).

#### 4.3.2 Flow Cytometry Data

For each method with usable code, the rules for the subpopulations were constructed on the Marburg and Dresden training data sets. Computations per method were limited to 72 hours. The accuracies of the classifiers were calculated on the sample data files using up to 50 rounds of cross-validation within the time limit.
Neither RF_LIME nor SuperFlowType was able to compute results on a personal computer (iMac PRO, 32 Cores, 256 GB RAM) in 72 hours of computing time. Therefore, to be able to compute results, the M32ms Microsoft Azure Cloud Computing system consisting of 32 Cores and 875GB RAM (9.180163 EUR/hour) was used. Still, neither the methods RF nor FlowType (of RF_LIME and SuperFlowType) were able to provide results of the full training set of the data. Thus, a 20% sample was used, which resulted in 40h computation time for RF_LIME and a 24h computation time for SuperFlowType per trial in the case of the Marburg data set. For the Dresden dataset, RF_LIME was stopped after 72 hours without a result. Computation of results for SuperFlowType retook 24hours with the procedure defined above.

## 5  RESULTS

The algorithms'' application to the Iris data with gaussian noise serves as a basic test of their performance of the algororiithms introduced in the prior sections because the datasets for human medical research are of limited size and not available in large quantities. In contrast Iris is a well-known biological dataset for which the performance of algorithms is expected to be high.The data comprises three distinct classes (k =3), which have to be diagnosed using the d = 4 variables. Table 1 shows that the accuracies of all algorithms besides SuperFlowType is above 90%. The largest difference is in the number of identified clusters and the number of conditions for the description of a cluster. However, all algorithms deliver sufficiently simple descriptions with typically less than 9 clusters. Suprisingly, the clusters delived by FAUST do not match the clusters of iris meaning that the descriptions are not useful.

The eUD3.5 algorithm could not be transferred to the Marburg and Dresden data set (see above). For each method with usable open-source code, the subpopulation'' rules were constructed on the two training data sets (details see below). Computations per method were limited to 72 hours. Accuracies of the resulting classifiers on the test data were calculated on the respective sets of patient's data files using up to 50 rounds of cross-validation within the time limit. Neither RF_LIME nor SuperFlowType was able to compute results on a personal computer (iMac PRO, 36 Cores, 256 GB RAM) in 72 hours of computing time. Therefore, to be able to compute results, the M32ms Microsoft Azure Cloud Computing system consisting of 36 Cores and 875GB RAM (9.180163 EUR/hour) was used. Still, neither the sub-methods RF nor flowType (of RF_LIME and SuperFlowType) provided results of the dat''s full training set. Thus, a 20% sample was used, which resulted in 40h computation time for RF_LIME and a 24h computation time for SuperFlowType in the case of the Marburg data set. Table 2 shows the results.

For the Dresden dataset, RF_LIME was stopped after 72 hours without a result. The computation of results for SuperFlowType took 24 hours again with the procedure defined above. Table 3 presents the results. Contrary to the compared algorithms, ALPODS finished after 1 minute CPU time on the iMac PRO, 32 Cores, 256 GB RAM) Therefore only ALPODS was able to perform the desired 50 cross-validations.

For the Marburg dataset, ALPODS identified five relevant populations for distinguishing bone marrow and peripheral blood. Using the visualization techniques, the clinical experts could understand the population and identify the following cell types in the population:

FlowType starts with all $3^d$ possible conditions in all d variables. For the flow cytometric data, this means more than 50.000 populations. Optimix reduced this to typically around 8000 populations, and the subsequent ABC analysis, which took the significance of the clusters for a diagnosis into account, reduced this to an average of 2744 populations. However, this number of clusters is too large to even look into for further explanations. Only a fraction of the results had an acceptable accuracy of greater than 80%. The number of clusters identified in RF_LIME was typically 20 to 40.

Contrary to the claim of the authors of FAUST, that "it has a general purpose and can be applied to any collection of related real-valued matrices one wishes to partition" [Greene et al., 2021], FAUST failed to reproduce the cluster structures the iris dataset (0.64 accuracy). As their intitial benchmarking was neither unbiased nor compared against state-of-the-art clustering algorithms (c.f. discussion in [Thrun, 2021]), it would require further studies to investigate if FAUST is able to explain given structures in data. Although for the flow cytometry datasets, the random forest classifier yielded a performance near equal to ALPODS, a large number of populations with eight conditions for each population are too complex to be understandable by domain experts without additional algorithms for selecting appropriate populations. Even then, describing each population by eight conditions for each condition containing up to three thresholds seems unfeasible. The results indicate that FAUST is probably not very useful to discovering knowledge in flow cytometric datasets. Moreover, the computation time rises with larger training datasets in comparison to ALPODS.

The results for the Dresden data set were similar. For the five clusters identified by ALPODS (See Table 4), the main cell types could be identified (See Table 5). Blood stem cells are located in the bone marrow niche and give rise to myeloid progenitor cells (population 1). At the developmental maturation stage of granulocytes these cells are released into the peripheral blood (population 2 and 4). T cells derive from the thymus and proliferate in the peripheral blood upon stimuli following infections (population 3). Therefore, these cell type predominantly occurs in the peripheral blood. Hematogones are B cell precursor cells which are found almost exclusively in the bone marrow (population 5). However, due to active cell trafficking the is no strict border between the two organic distributional spaces.

We further sought to verify the principle of the algorithm with another data set and compared BM samples diagnosed as normal (healthy) bone marrow (n=25) and samples with infiltration of leukemia cells (n=25). Within a computation time of minutes, ALPODS identified a major dominant population labeled "blast cells" by the flow cytometry experts in the VisPanels of Fig.3 vs. Fig.4 The population is described by one explanation resulting in an 98% classification accuracy. The blast cell population is characterized by high expression of the antigens CD34 and CD117, while CD45 is only moderately expressed as depicted in a conventional scatter plot (Figure 4). "Blast cell" is another term for "leukemia cell". A blast cell infiltration level of >20% defines the disease "leukemia" that is diagnosed out of bone marrow specimen by physicians. If a second population (magenta in VisPanels Fig, 3 and Fig. 4) defined by an additional explanation is taken into account, the accuracy on the test data is 100%. The comparable approach FAUST[Greene et al., 2021] identified 212 relevant populations after a computation time of 3 hours. The application of random forests on these populations resulted in an accuracy of 92% on the test data.

**Table 1.** Average running time, number of clusters, and number of conditions for explanation rules for the XAI algorithms on the Iris data set. The baseline of CART[Breiman et al., 1984] had an accuracy of 98% and required 4 rules. CART was computed with the R package "rpart" available on CRAN ""ttps://CRAN.R-project.org/package=rpart"). * For FAUST a classifier could not be learned, hence, the clustering accuracy between the given iris clusters and the FAUST generated clusters was measured by the procedure described in the R package "FCPS" on CRAN [Thrun/Stier, 2021].

| IRIS | eUD3.5 | FAUST* | RF-LIME | SuperFlowType | ALPODS |
|---|---|---|---|---|---|
| Processing Time | <1min | <1min | <1min | <1min | <1min |
| No of Crossvalidations | 50 | * | 50 | 50 | 50 |
| Max No Of Cluster | 15 | 4 | 14 | 30 | 5 |
| Mean No Of Cluster | 8+-3.4 | 4 | 6+-3.2 | 7+-3 | 2+-1 |
| Max No Of Conditions for a Cluster | 4 | 4 | 2 | 4 | 4 |
| Mean No Of Conditions for a Cluster | 2 +- 1.1 | 4 | 2 +- 0.5 | 3 +- 0.8 | 2 +- 0.7 |
| Performance Accuracy | 98 +- 0.5 | 64* | 96.3 +- 1.4 | 80 +- 10 | 96 +- 0 |

**Table 2.** Average running time, number of clusters, and number of conditions for explanation rules for the XAI algorithms on the Marburg data set. The baseline of CART[Breiman et al., 1984] had an accuracy of 50% and required 6 rules. CART was computed with the R package "rpart" available on CRAN ""ttps://CRAN.R-project.org/package=rpart").* Leave-one-out cross-validation.

| Marburg Dataset | RF-FAUST | RF-LIME | SuperFlowType | ALPODS |
|---|---|---|---|---|
| Processing Time | <1min | 72h | 24h | 1min |
| No of Crossvalidations | * | 2 | 3 | 50 |
| Max No Of Cluster | 79 | 40 | 5486 | 5 |
| Mean No Of Cluster | 79 | 21 +- 11.6 | 2744 +- 1583 | 3 +- 1 |
| Max No Of Conditions for a Cluster | 8 | 2 | 10 | 6 |
| Mean No Of Conditions for a Cluster | 8 | 2 +- 0.5 | 10 | 5 +- 0.8 |
| Performance Accuracy | 0.92 | 80.0 +- 0.0 | 71.6 +- 15.9 | 96.9 +- 0.9 |

**Table 3.** Average running time, number of clusters, and number of conditions for explanation rules for the XAI algorithms on the Dresden data set. The baseline of CART[Breiman et al., 1984] had an accuracy of 46% and required 9 rules. CART was computed with the R package "rpart" available on CRAN ""ttps://CRAN.R-project.org/package=rpart") .

| Dresden Dataset | RF-FAUST | RF-LIME | SuperFlowType | ALPODS |
|---|---|---|---|---|
| Processing Time | 17 min | >72h | 24h | 1min |
| No of Crossvalidations | 50 | - | 1 | 50 |
| Max No Of Cluster | 43 | - | 1456 | 5 |
| Mean No Of Cluster | 43 | - | 2744 +- 1583 | 3 +- 1 |
| Max No Of Conditions for a Cluster | 8 | - | 10 | 6 |
| Mean No Of Conditions for a Cluster | 8 | - | 10 | 4 +- 1.1 |
| Performance Accuracy | 96 +- 0.06 | - | 70 | 96.8 +- 0.9 |

**Table 4.** Description rules generated by ALPODS for Marburg dataset, which could be described by human experts and their occurrence frequencies in peripheral blood and bone marrow.

| Pop No | CellTypes | Description Rule | Frequencies in [%] Peripheral Blood | Bone Marrow |
|---|---|---|---|---|
| 1 | Myeloid progenitor cells | SS++, CD33+, CD13- | 5.0 | 43.0 |
| 2 | Subcellular events and aggregates | SS--,HLA_DR--,CD45-, CD117not+, CD33- | 19.0 | 1.0 |
| 3 | Progenitor B-cells | SS-, HLA_DR+, CD4 not++ | 0.3 | 2.0 |
| 4 | Thrombocyte aggregations | SS-,HLA_DR--,CD13-, CD117not(-),CD34not(--), CD117not(+) | 4.2 | 0.5 |
| 5 | CD34 positive early progenitor Cells | SS0, CD45-, CD34+ | 1.0 | 8.0 |

**Table 5.** Description rules generated by ALPODS for Dresden dataset, which could be described by human experts and their occurrence frequencies in peripheral blood and bone marrow.

| Pop No | CellTypes | Description Rule | Frequencies in [%] Peripheral Blood | Bone Marrow |
|---|---|---|---|---|
| 1 | Myeloid progenitor cells | FS+, CD45-, CD13- | 2.0 | 26.9 |
| 2 | Mature granulocyte subset | SS++,CD7-not(--), CD117+, CD13+not(++) | 13.0 | 2.9 |
| 3 | T-cells | SS-not(--), below average CD33, CD45+, CD13- | 12.2 | 2.1 |
| 4 | Granulocytes subset | Below average HLA_DR, below average CD33, above average CD7, CD117-, SS+, above average CD13 | 9.1 | 1.9 |
| 5 | Hematogenes with lymphocyte subset | CD33-, FS -, not(CD45+), CD13- | 2.6 | 8.6 |

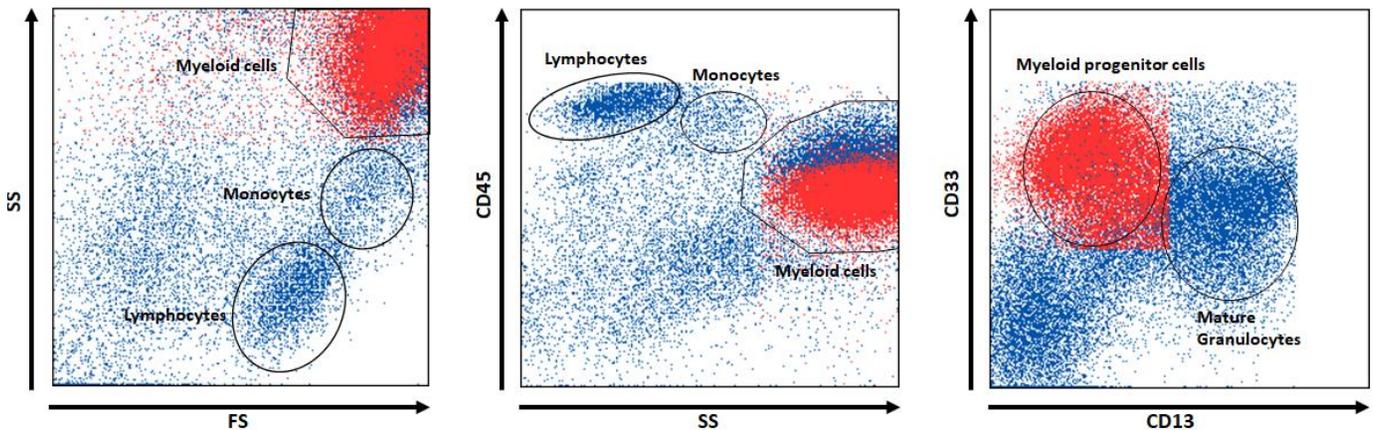

**Figure 1** A panel of flow cytometry dot (scatter) plots of events from the training set, i.e. a composite of all cases. Red dots denote subpopulation one, which was recognized by ALPODS as relevant for bone marrow identification.

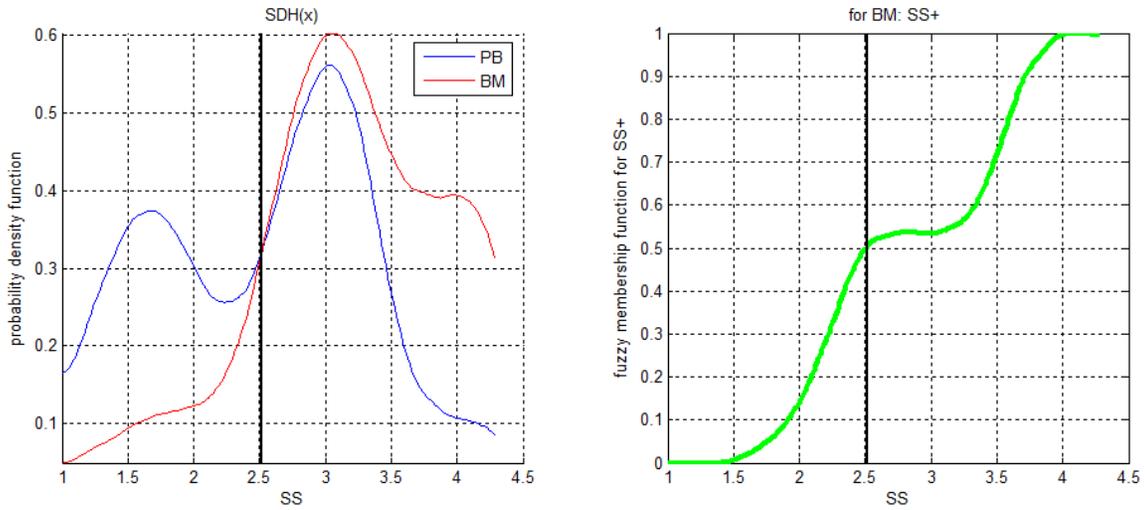

**Figure 2** Probability density functions (left) and derived fuzzy membership function for "SS+ in BM"(right)

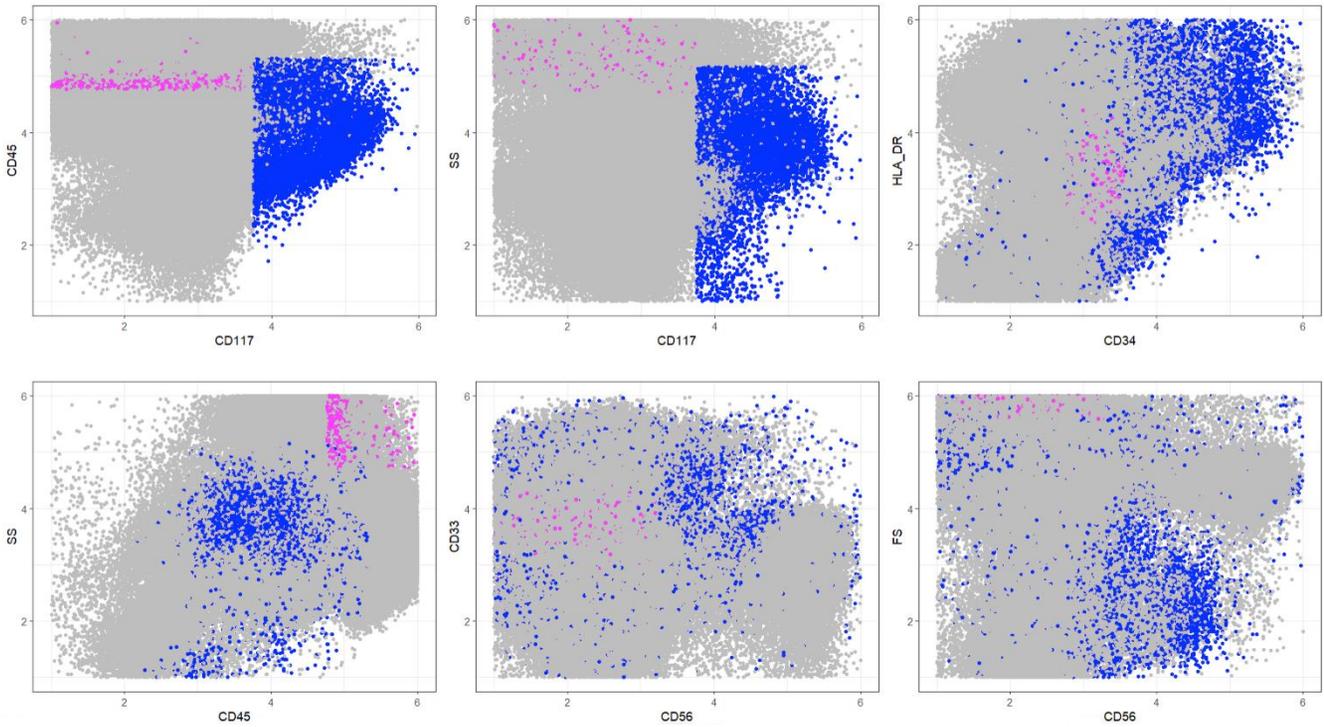

**Figure 3** VisPanel of flow cytometry dot (scatter) plots of events from the training set, i.e. a composite of all cases. Red dots denote subpopulation one, which was recognized by ALPODS as relevant for leukemia identification in a healthy bone marrow sample.

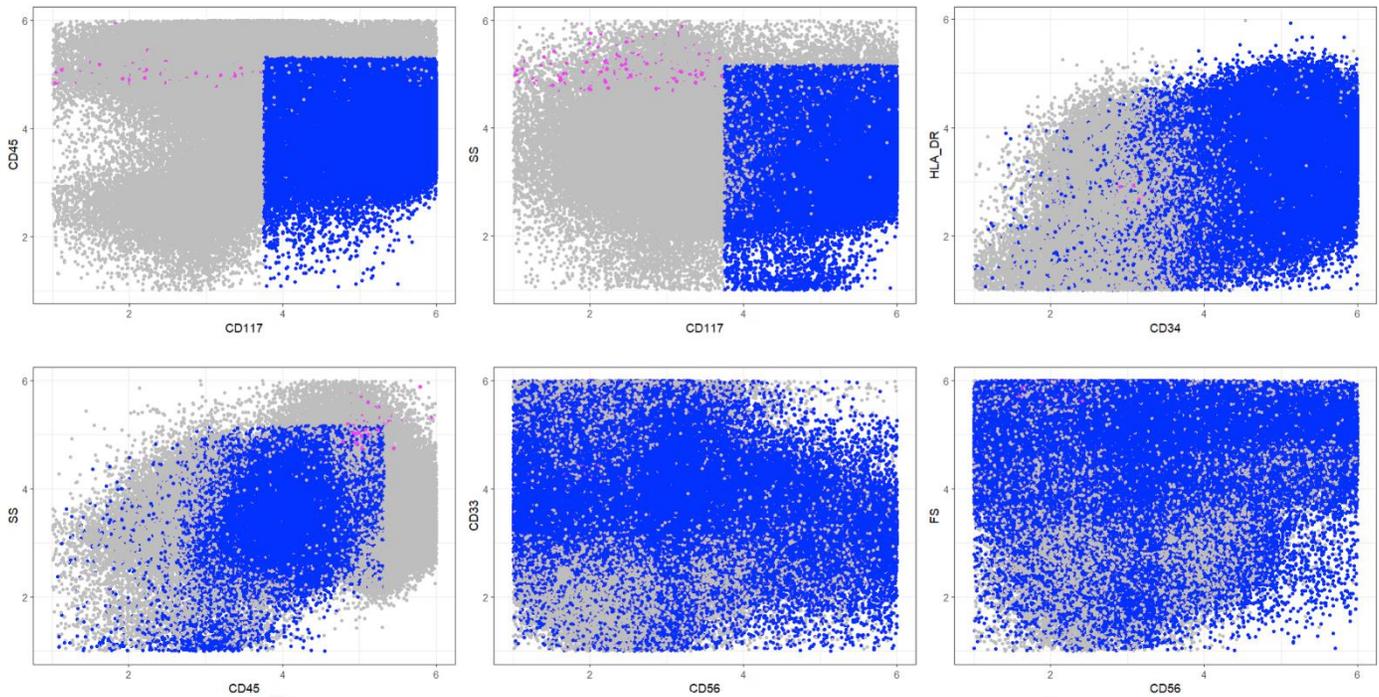

**Figure 4** VisPanel of flow cytometry dot (scatter) plots of events from the training set, i.e. a composite of all cases. Red dots denote subpopulation one, which was recognized by ALPODS as relevant for leukemia identification in an bone marrow sample of a patients having AML.

## 6    DISCUSSION AND CONCLUSION

Artificial intelligence systems constructed by machine learning have shown in recent years that these algorithms are able to perform classification tasks with an precision similar to experts in a domain. Subsymbolic systems, as, for example, artificial neural networks, are usually the best performing systems.  However, the subsymbolic approach sacrifices deliberately explainability, i.e., human understanding, as a tradeoff for performance. For life-critical applications, such as medical diagnosis, artificial intelligence systems are required, which are able to explain their decisions (explainable AI systems). Here explainable AI systems (XAI) are considered, which identify clusters (subpopulations) within a dataset that are first relevant for the diagnosis and second, describable in the form of a set of conditions (rule) that are in principle human-understandable and genuinely reflect the clinical condition. For such systems, several criteria are relevant. From a practical standpoint, the algorithm should be able to finish its task in a reasonable time on typical hardware. Second, the performance in terms of decision accuracy should exceed 80-90%, and third, and most importantly, the results must be understandable in the sense that they are interpretable by and explainable to human domain experts.

In this work, we introduced an XAI algorithm (ALPODS) designed for large (n ≥ 100.000) and multivariate data (d≥10), such as typical flow cytometry data. The distinction between bone marrow and peripheral blood serves as an example because many approaches have been proposed to solve the problem of bone marrow dilution with peripheral blood [Holdrinet et al., 1980; Delgado et al., 2017]. Holdrinet et al. suggested counting of nucleated cells by an external hemacytometer device while others suggest to run a separate flow cytometric panel to quantify those populations that predominantly occur in the BM or PB[Abrahamsen et al., 1995; Delgado et al., 2017]. Of note, the ALPODS algorithm is capable to extract information about population differences from both datasets thus making additional external or additional measurements dispensable. In comparison to other state of the art XAI algorithms for the same task, ALPODS delivered human-understandable results. Experts in flow cytometry identified the cell populations as "true" progenitor cells, T cells, or granulocytes. Moreover, experts verified that the distribution pattern of these populations in PB or BM is plausible according to the knowledge of hematologists. To enhance the understandability, a visualization tool relying on differential density estimation between the cases in a cluster vs. the rest of the data was essential because it seems that in several domains visualizations improve the understandability of explanations significantly. For example, in xDNN, appropriate training images are algorithmically selected as prototypes and combined into logical rules in the last layer called mega clouds. In order to explain the decision of the CNN[Angelov/Soares, 2020]. Similar to the visualization panel (VisPanel) of the here introduced method ALPODS, the presented rules are human-understandable because they are presented as images. However, the generalization of the explanation process is the neural networks well-known application of a Voronoi-tessellation [Lötsch/Ultsch, 2014; Thrun/Ultsch, 2020c] of the projection of principal component analysis (PCA) [Hotelling, 1933]. If a high-dimensional dataset, for example N>18.000 gene expressions of patients diagnosed with a variety of cancer illness is considered [Thrun, 2021], neither the logical rules consisting of prototypes nor the Voronoi-tessellation of PCA would be understandable to a domain expert. Moreover, it is well-known that PCA does not reproduce high-dimensional structures appropriately [Thrun/Ultsch, 2020c] which could lead to untrustworthy explanations.

The explainable AI expert system using the results of ALPODS is able to correctly distinguish between two material sources by identifying cellular and subcellular events, i.e., bone marrow and peripheral blood. In particular, every diagnosis by this XAI system can be understood and validated by human experts. Moreover, in line with the argumentation taken by Rudin [Rudin, 2019], are results indicated that a trade-off between understandability and accuracy is not necessary for XAIs that use biomedical data.

The differentiation between peripheral blood and bone marrow is manageable for human experts in flow cytometry without AI. Therefore, this problem is perfectly suited as a proof of concept for an XAI because it is comprehensible and verifiable to human experts, as we showed here.

However, the cell population-based approach of ALPODS is universally applicable for high-dimensional biomedical and other data and may help answer meaningful questions in the diagnostics and therapy of cancer, hematological blood- and bone marrow diseases. In practice, ALPODS may facilitate the sophisticated differential diagnosis of lymphomas and may help to identify prognostic subgroups. Most importantly, ALPODS reduces the complex high dimensional flow cytometry data to the essential disjunct cell populations. Only this essential information is usable for humans and could assist clinical decisions in the future. One additional advantage of the algorithm is that its computation time is significantly faster than comparable algorithms' computation time. It should be noted that the intended usage of ALPODS are datasets in which each case itself comprises of many events. In contrast, many typical explainable AI methods focus on data for which a case is stored within one event described by a set of features.

## SUPPLEMENTARY A: APPLICATION DETAILS FOR RF-LIME AND RF-FAUST

To compute RF-LIME the R package "lime published on CRAN and the R package "randomForestSRC" published on CRAN were used (https://CRAN.R-project.org/package=randomForestSRC and " https://CRAN.R-project.org/package=lime). The package "randomForestSRC is a fast OpenMP parallel computing of Breiman's random forests. The R package 'lime' is a port of the 'lime' 'Python' package.

The functionality of the package "lime" was applied in the default parameter setting. LIME explanation are based on a linear model with LASSO regularization. Although the value of penalty should influence the number of clusters in Table 1-3, it was unfeasible to change this value in the R package due to a lack of documentation. Hence, this hypothesis could not be investigated further.

The computation random forests performed swiftly for the datasets. However, the computation of "lime" took a long time. We evaluated this explicitly with profiling (utils::Rprof) in the case of N=1500 of iris data with gaussian noise which yielded an overall computation time of 118 seconds of which the function lime::explain tool 114 seconds itself. In Figure 5 the facetted heatmap of RF-LIME is shown of the LIME algorithm [Ribeiro et al., 2016] using the R package "lime".

Faust (https://github.com/RGLab/FAUST) provides cell-counts per populations as an output. Each population in FAUST is described in form of a set of CDi+, CDj- conditions. Greene et al., provided several unsupervised and one supervised application based on a linear regression model for FAUST in their work [Greene et al., 2021]. Here, the supervised application is adapted by using FAUST, as is proposed by the authors, as a feature extraction method and combining it with a typical random forest classifier because random forests are claimed to have the best performance [Fernández Delgado et al., 2014].

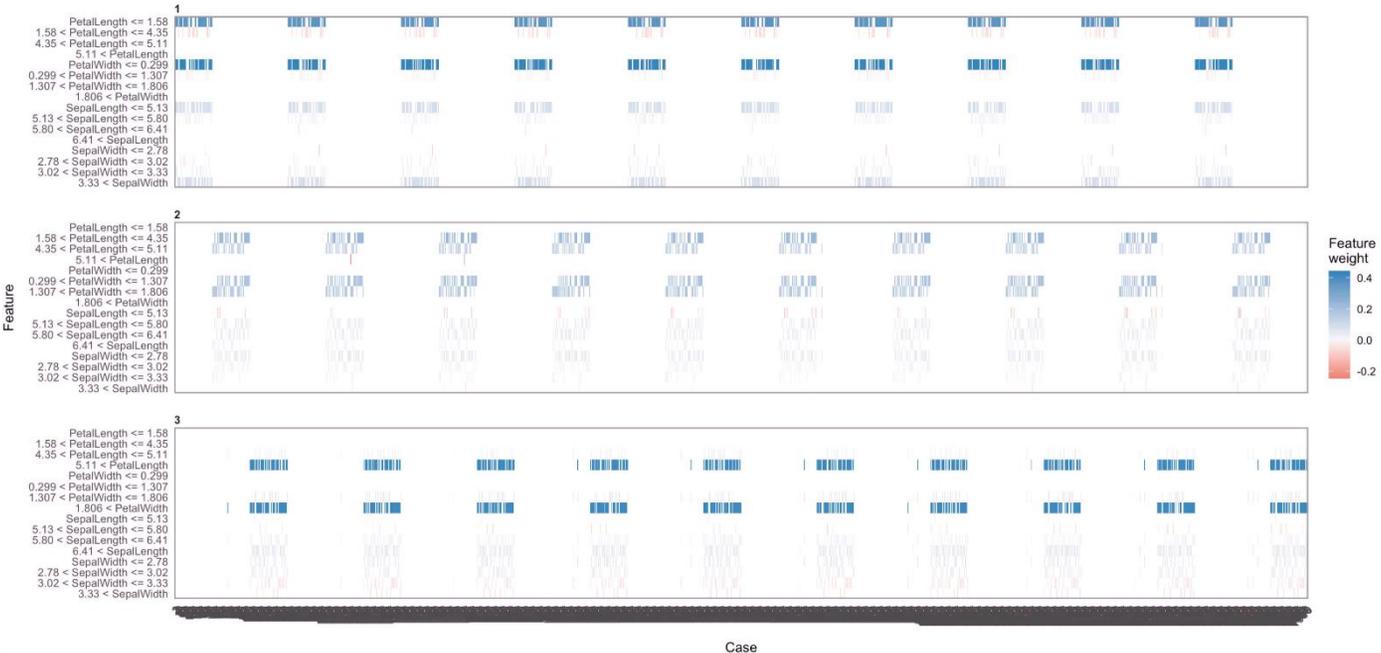

**Figure 5** Facetted heatmap of RF-LIME for the Iris data.

*Acknowledgements*

We thank Quirin Stier for the application of the FAUST algorithm to the training sets and the extraction of the rules from these training sets.